\documentclass[11pt]{article}

\usepackage[]{acl}

\usepackage{times}
\usepackage{latexsym}
\usepackage{multirow}
\usepackage{graphicx}
\usepackage{graphicx} 
\usepackage{caption}
\usepackage{subcaption}
\usepackage{xcolor}
\usepackage{colortbl}

\usepackage[T1]{fontenc}

\usepackage[utf8]{inputenc}

\usepackage{microtype}

\title{WADER at SemEval-2023 Task 9: A Weak-labelling framework for Data augmentation in tExt Regression Tasks}

\author{Manan Suri\textsuperscript{\rm{1}}\thanks{*Equal contribution.}, Aaryak Garg\textsuperscript{\rm{1}*}, Divya Chaudhary\textsuperscript{\rm{2}}, \\
{\bf Ian Gorton\textsuperscript{\rm{2}},} {\bf Bijendra Kumar\textsuperscript{\rm{1}}}\\
 Netaji Subhas University of Technology, New Delhi\textsuperscript{\rm{1}} \\
 Northeastern University, Seattle \textsuperscript{\rm{2}} \\
\texttt{\{manan.suri.ug20,aaryak.ug20\}@nsut.ac.in}
}

\begin{document}
\maketitle
\begin{abstract}
Intimacy is an essential element of human relationships and language is a crucial means of conveying it. Textual intimacy analysis can reveal social norms in different contexts and serve as a benchmark for testing computational models' ability to understand social information. In this paper, we propose a novel weak-labeling strategy for data augmentation in text regression tasks called WADER. WADER uses data augmentation to address the problems of data imbalance and data scarcity and provides a method for data augmentation in cross-lingual, zero-shot tasks. We benchmark the performance of State-of-the-Art pre-trained multilingual language models using WADER and analyze the use of sampling techniques to mitigate bias in data and optimally select augmentation candidates. Our results show that WADER outperforms the baseline model and provides a direction for mitigating data imbalance and scarcity in text regression tasks.
\end{abstract}

\section{Introduction}

Intimacy is considered a fundamental element of human relationships, as recognized by several scholars \cite{sullivan,maslow,prager}. Research indicates that intimacy can be modeled computationally and that textual intimacy is a crucial aspect of language \cite{pei2020quantifying}. Analyzing textual intimacy can reveal social norms in various contexts and serve as a benchmark to test computational models' ability to understand social information \cite{pei2020quantifying, hovy-yang-2021-importance}. Moreover, intimacy plays a critical role in human development and well-being \cite{zimmerman,sneed}, and language is an essential means of conveying it in a social context. Individuals negotiate intimacy in language to fulfill fundamental and strategic needs while respecting social norms. Task 9 of SemEval 2023 \cite{pei2022semeval} aims to quantify intimacy in a multilingual context, with evaluation on tweets from 10 languages. The training corpus for the task consists of tweets in English, Spanish, Italian, Portuguese, French, and Chinese. The testing corpus additionally contains tweets from Hindi, Arabic, Dutch and Korean. 

The novelty of our strategy, WADER (Weak-labeling strategy for Data augmentation in tExt Regression Tasks) is the use of data augmentation to A) solve the problem of an imbalance distribution of data, B) augment data for a cross-lingual zero-shot set-up. WADER uses the distribution to selectively sample texts with lower representation in the label distribution, uses translation to augment sentences and validates the augmentations against a baseline model, using a distribution based sampling approach. We finetune State-of-the-Art pre-trained language models including XLM RoBERTa \cite{xlmr} and XLNET \cite{xlnet}. Real world datasets are plagued by the problems of data imbalance and data scarcity, and WADER provides a direction for mitigating these problems for text regression tasks. WADER ranks 32nd overall across languages, 34th on seen languages and 29th on unseen languages. Our code has been released on GitHub. \footnote{Code will be released on paper acceptance.}

The main contributions of this paper are as follows:
\begin{enumerate}
    \item Provide a data augmentation framework specific to text regression.
    \item Provide a method for data augmentation in cross-lingual, zero-shot tasks.
    \item Benchmark performance of pre-trained language models.
    \item Analysis of use of sampling techniques to mitigate bias in data and optimally select augmentation candidates.
\end{enumerate}

\begin{table*}[h]
\centering

\begin{tabular}{lrrrrrr}
\hline
\textbf{Language}   & \multicolumn{1}{l}{\textbf{Count}} & \multicolumn{1}{l}{\textbf{Mean}} & \multicolumn{1}{l}{\textbf{Std. Dev.}} & \multicolumn{1}{l}{\textbf{25th \%ile}} & \multicolumn{1}{l}{\textbf{50th \%ile}} & \multicolumn{1}{l}{\textbf{75th \%ile}} \\ \hline
\textbf{English}    & 1587                               & 1.89                              & 0.877273                               & 1.2                                     & 1.6                                     & 2.4                                     \\
\textbf{Chinese}    & 1596                               & 2.27                              & 0.93851                                & 1.5                                     & 2                                       & 2.8                                     \\
\textbf{French}     & 1588                               & 2.06                              & 0.886265                               & 1.34                                    & 2                                       & 2.6                                     \\
\textbf{Italian}    & 1532                               & 1.94                              & 0.835105                               & 1.25                                    & 1.8                                     & 2.425                                   \\
\textbf{Spanish}    & 1592                               & 2.21                              & 0.941339                               & 1.4                                     & 2                                       & 2.8                                     \\
\textbf{Portuguese} & 1596                               & 2.16                              & 0.872903                               & 1.4                                     & 2                                       & 2.8                                     \\
\textbf{Overall}    & 9491                               & 2.09                              & 0.903512                               & 1.4                                     & 2                                       & 2.67                                    \\ \hline
\end{tabular}%
\caption{Description of the training set.}
\label{tab:1}
\end{table*}

\begin{figure*}[h!]
\centering
\begin{subfigure}{0.3\textwidth}
\includegraphics[width=\textwidth]{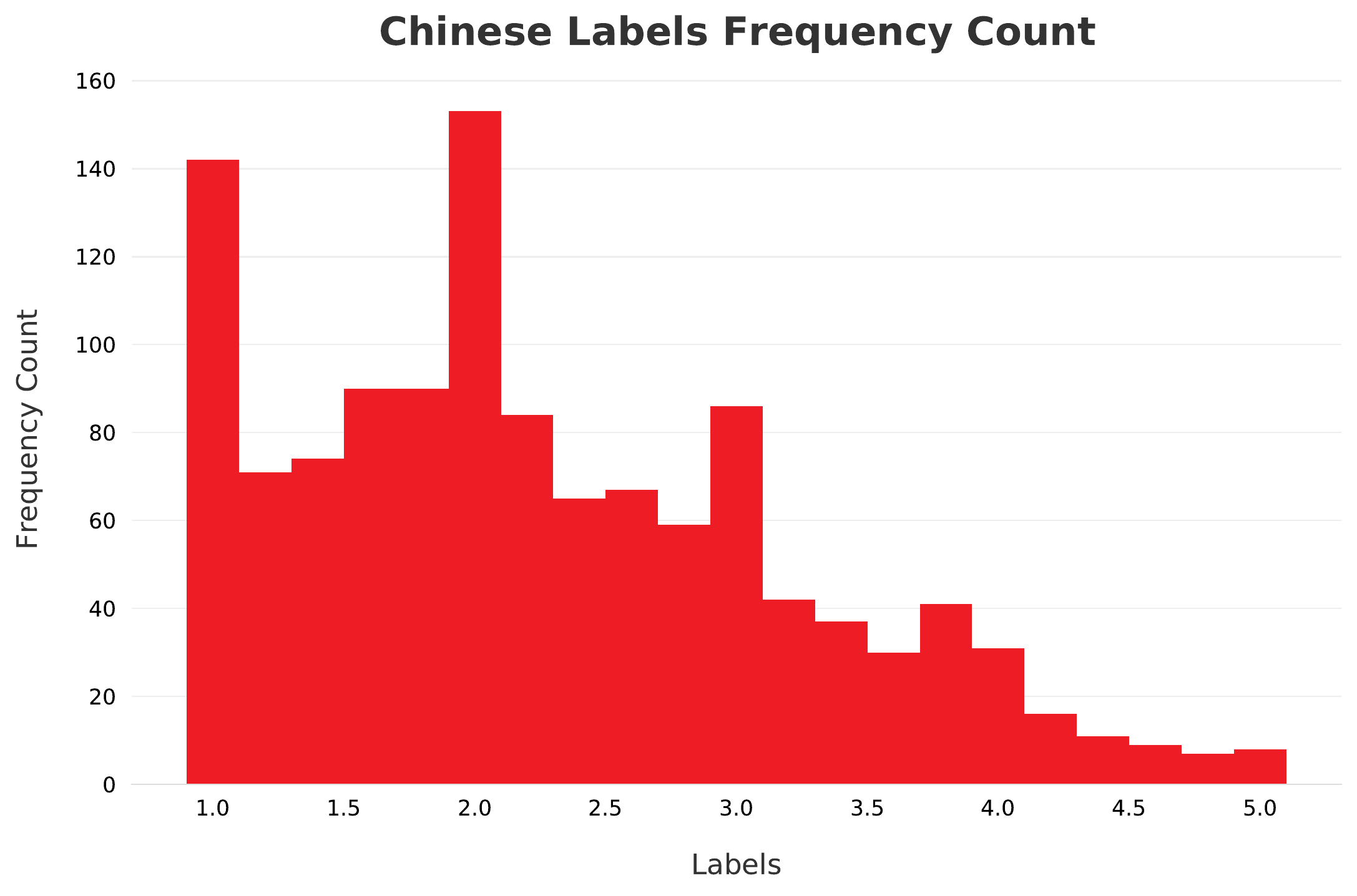}
\caption{ Chinese }
\label{fig:zh-hist}
\end{subfigure}
\begin{subfigure}{0.3\textwidth}
\includegraphics[width=\textwidth]{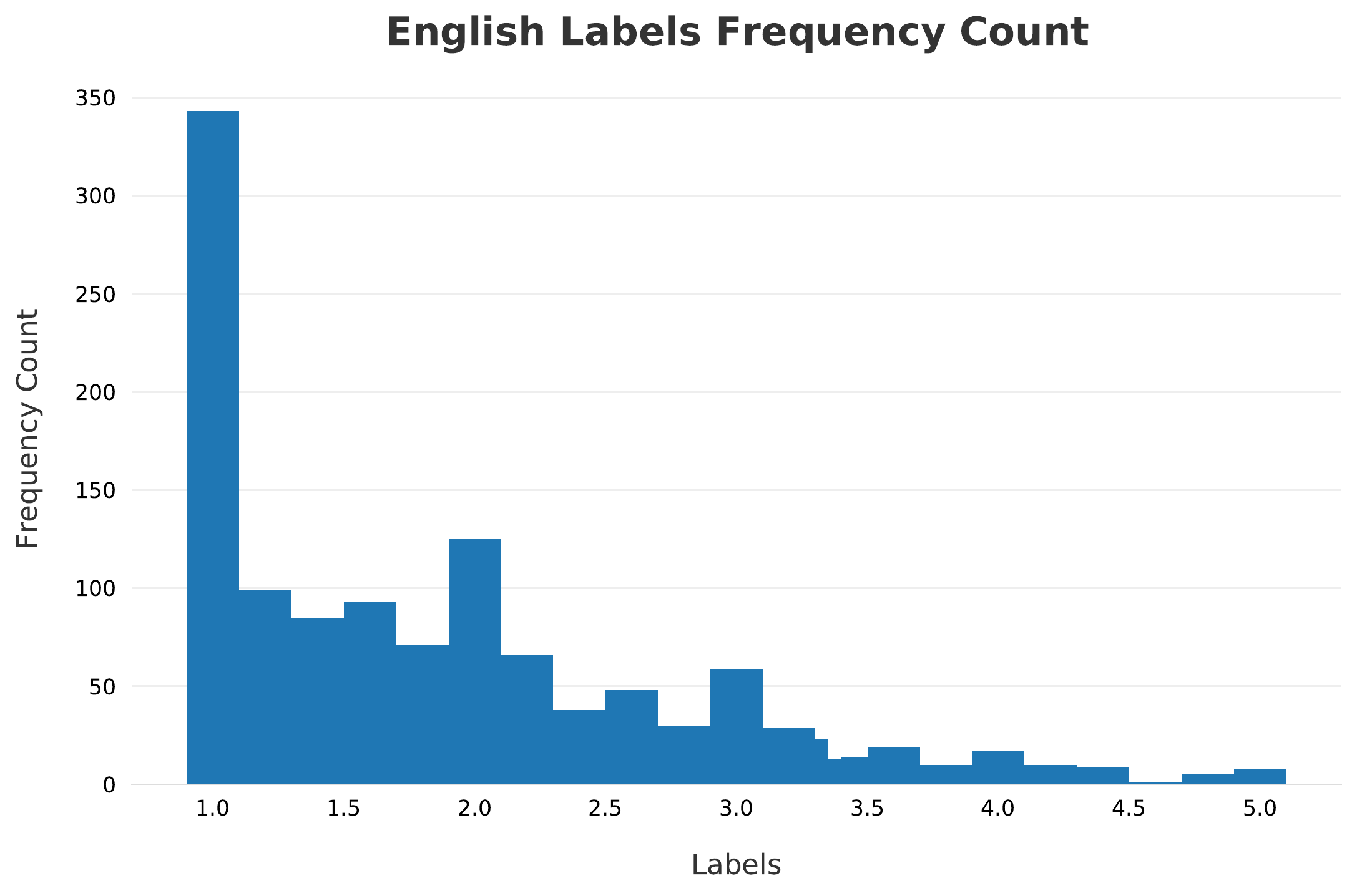}
\caption{English}
\label{fig:en-hist}
\end{subfigure}
\begin{subfigure}{0.3\textwidth}
\includegraphics[width=\textwidth]{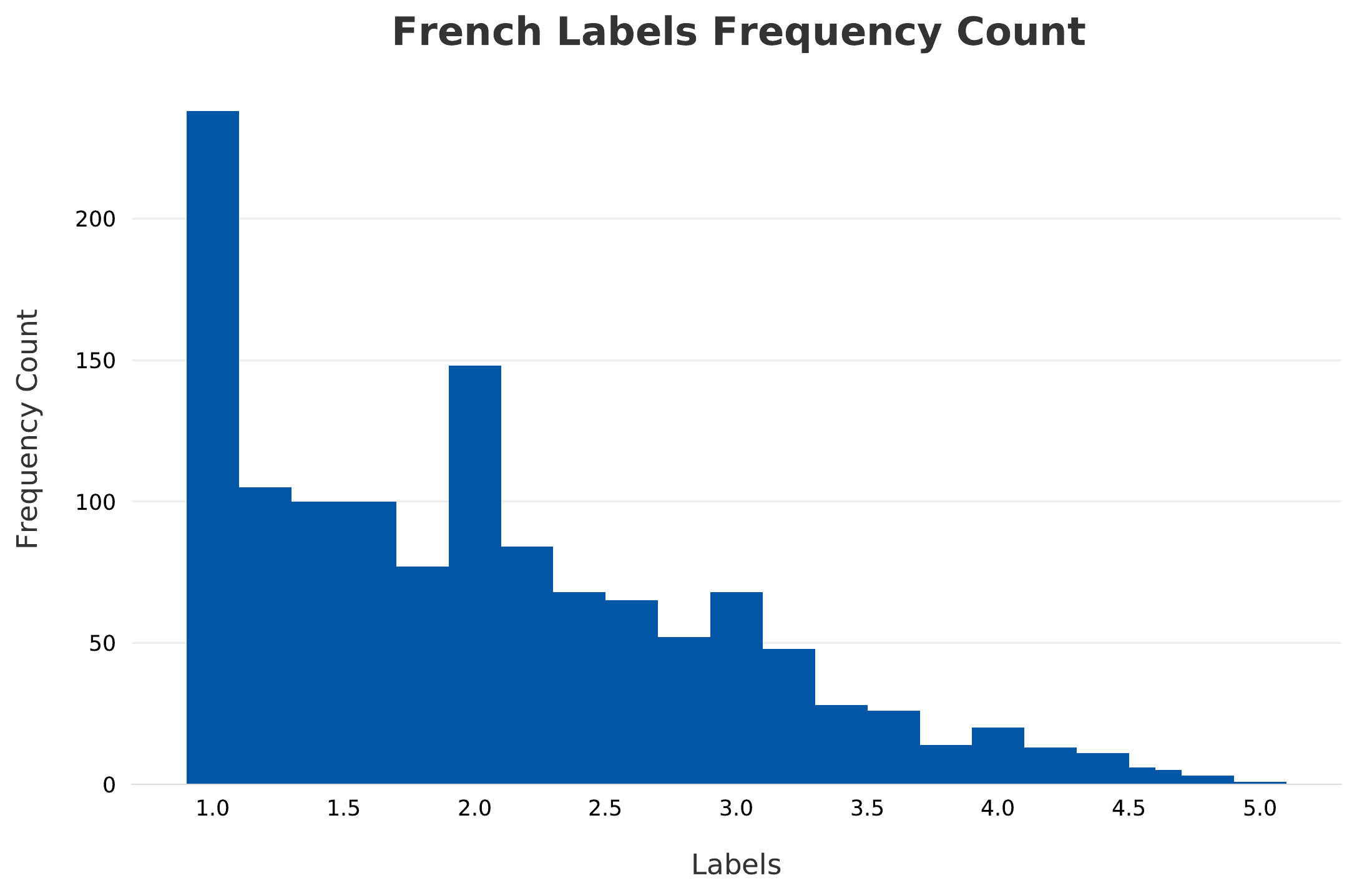}
\caption{French}
\label{fig:fr-hist}
\end{subfigure}

\begin{subfigure}{0.3\textwidth}
\includegraphics[width=\textwidth]{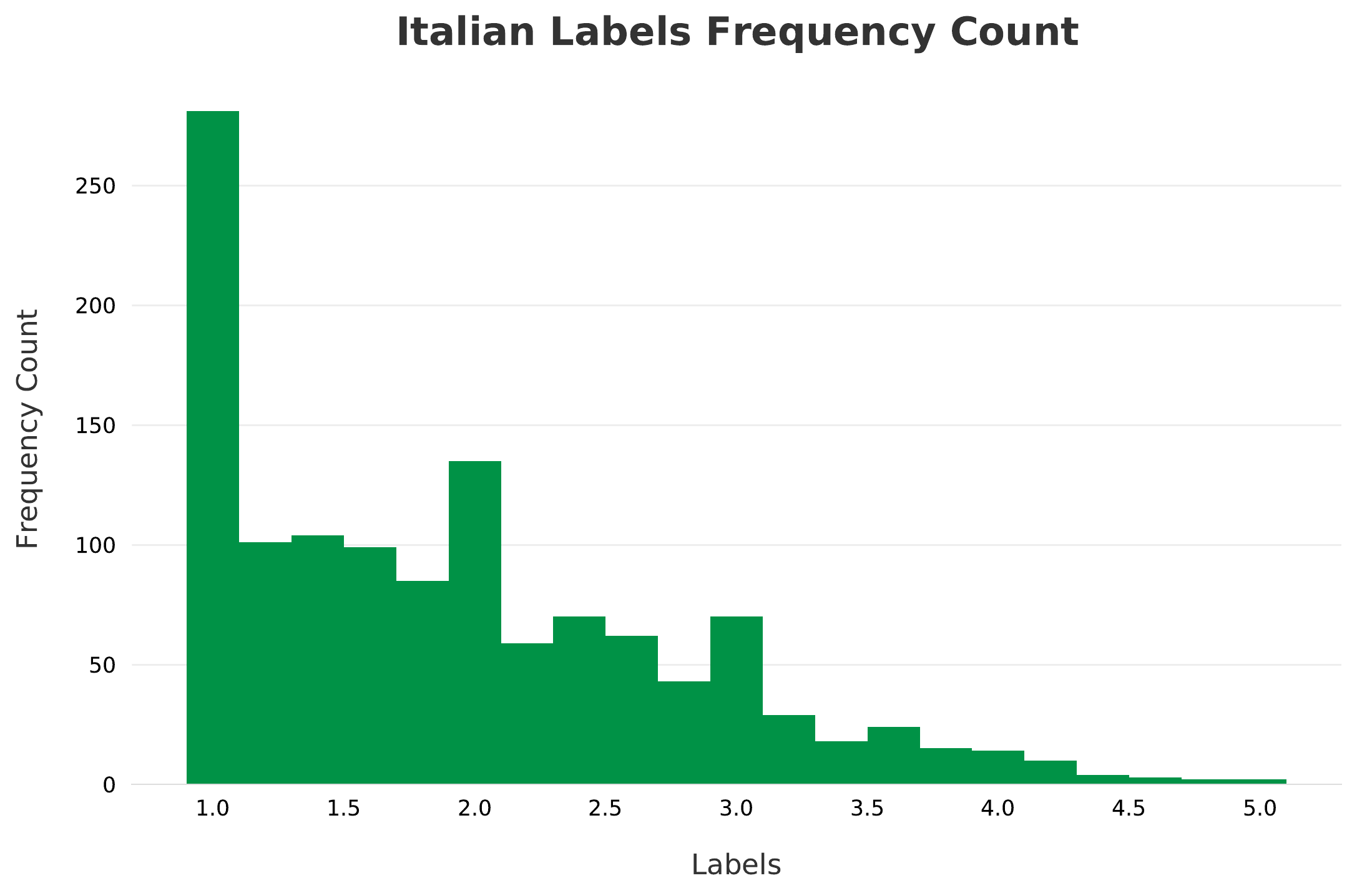}
\caption{Italian}
\label{fig:it-hist}
\end{subfigure}
\begin{subfigure}{0.3\textwidth}
\includegraphics[width=\textwidth]{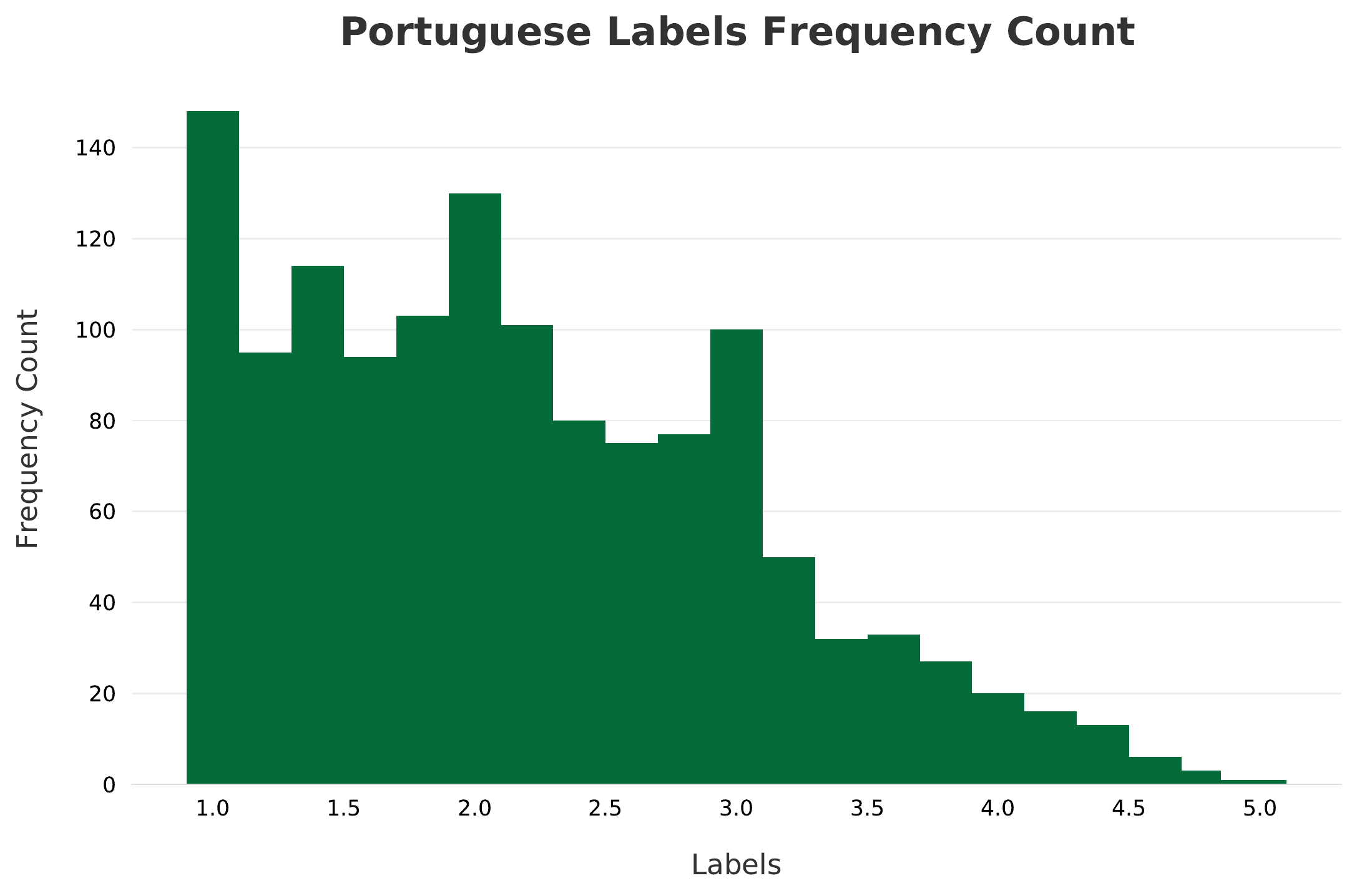}
\caption{Portuguese}
\label{fig:pt-hist}
\end{subfigure}
\begin{subfigure}{0.3\textwidth}
\includegraphics[width=\textwidth]{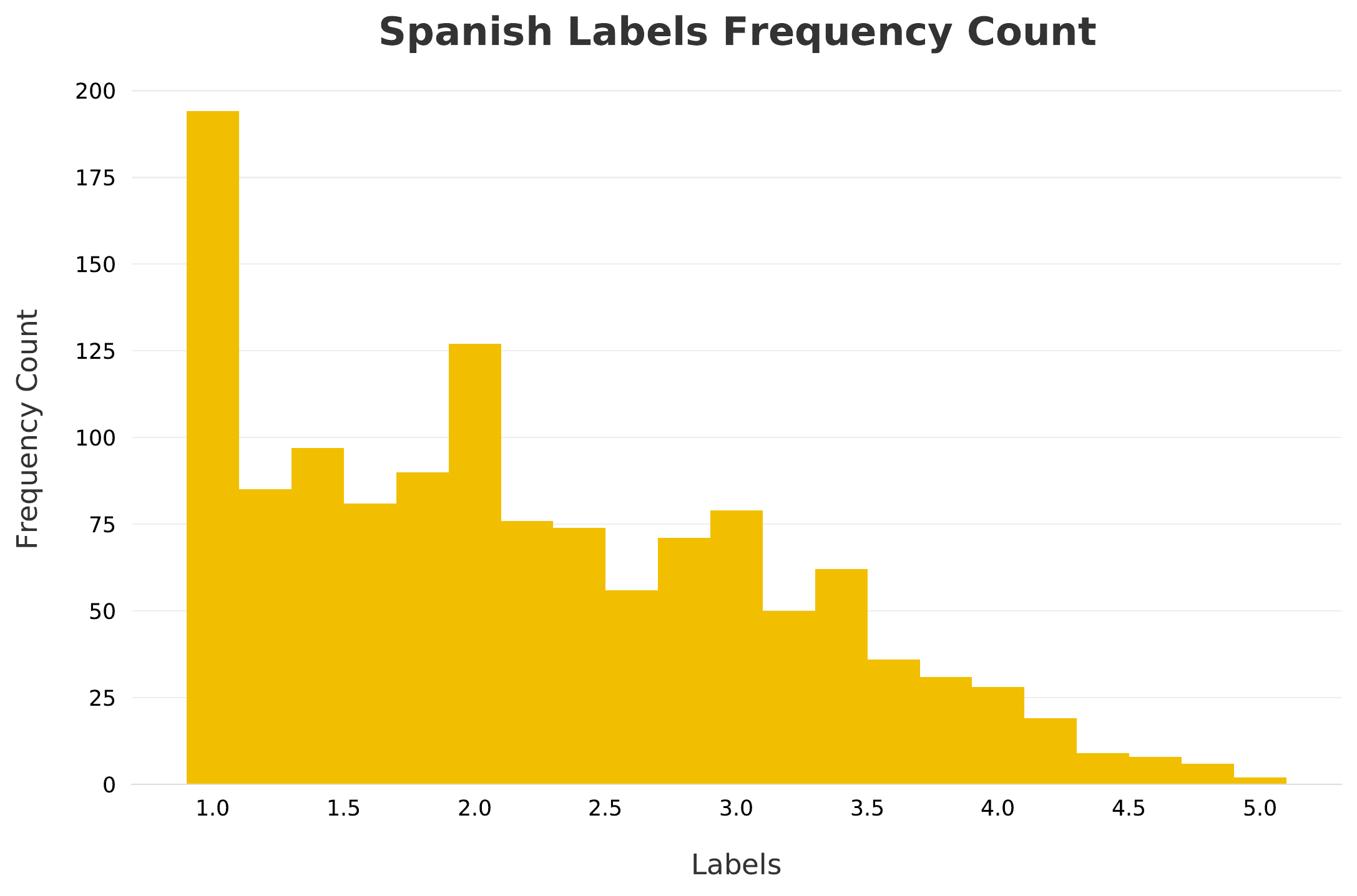}
\caption{Spanish}
\label{fig:es-hist}
\end{subfigure}

\caption{Frequency plots for different languages in the training set.}
\label{fig:overall}
\end{figure*}

The paper is organized as follows: Section 2 provides background information on the research, including a review of relevant literature, details about the task at hand, and information on the data used. Section 3 presents an overview of our approach, followed by a discussion of the experimental set-up in Section 4. The results of our study are analyzed in Section 5, and the paper concludes with a summary of findings and future directions for research in Section 6.


\section{Background}

\subsection{Past Work}

Data imbalance and scarcity are problems that are rampant in real world datasets. The high cost of obtaining large amounts of data, and expert annotations, a wealth of research has been done to support limited data settings. Data augmentation for text is broadly done in two ways, conditional data augmentation which involves data augmentation conditioned by the target label, and unconditional data augmentation which involves working with the corpus features only \cite{aug1,aug2}. Conditional data augmentation is done usually by deep generative models and pre-trained language models such as BART \cite{BART}, CBERT \cite{cbert}, GPT2 \cite{gpt2}. Common ways to perform unconditional data augmentation are lexical substitution and back translation. \citep{EDA} introduce several lexical techniques to augment textual data, including synonym replacement, random insertion, random swap and random deletion. However, these methods suffer from lack of sufficient diversity and often produce sentences that are not coherent. Back-translation especially has received widespread attention, because progress in machine translation has made back-translation an efficient way to generate diverse sentences in the dataset without compromise in coherence and semantic quality. Common translation tools used are seq2seq based models, NMT and transformers. Different techniques exist for text classification and NER tasks, but to the best of our knowledge our work is unique in the text regression domain.

Weak supervision of text labeling during data augmentation is an example of Semi-Supervised Learning (SSL) methods. The main idea of these methods is to regularize the learning process by  training a network with the given data, using the network to label unlabelled data and finally use both the true-labeled and weak-labeled data points to train the final model.

\subsection{Task Description}

SemEval 2023 Task 9: Multilingual Tweet Intimacy Analysis \cite{pei2022semeval} is a task that deals with detecting intimacy in 10 languages. This task is co-organized by University of Michigan and Snap Inc. Intimacy is a fundamental aspect of human relationships, and studying intimacy in  a textual context has many potential applications in the field of computational linguistics.  The training data is available in 6 languages: English, Spanish, Italian, Portuguese, French, and Chinese. The evaluation is done on the given training languages, as well as 6 unseen languages: Hindi, Arabic, Dutch and Korean.

The metric of evaluation for the task is Pearson’s R. Pearson’s R, $r$ is expressed as follows for two variables $x$ and $y$:
\begin{equation}
   r=\frac{\sum\left(x_i-\bar{x}\right)\left(y_i-\bar{y}\right)}{\sqrt{\sum\left(x_i-\bar{x}\right)^2 \sum\left(y_i-\bar{y}\right)^2}} 
\end{equation}

The correlation coefficient $r$ ranges from -1 to 1, with an absolute value of 1 indicating a perfect linear relationship between the two variables. In such a case, all data points lie on a line that can be represented by a linear equation. The sign of the correlation coefficient is determined by the regression slope, with a value of +1 indicating that all data points lie on a line where $y$ increases as $x$ increases, and a value of -1 indicating the opposite. A correlation coefficient of 0 implies that there is no linear dependency between the two variables.

\subsection{Data Description}
\label{sec:data}
The dataset for the task is the MINT- Multilingual INTimacy analysis \cite{pei2022semeval} dataset. The training set contains sentences in 6 languages: Chinese, English, French, Portuguese, Spanish and Italian. The dataset has 9491 tweets. Distribution of sentences in different languages is given in Table \ref{tab:1}.

Intimacy is annotated using a 5-point Likert scale where 1 indicates “Not intimate at all” and 5 indicates “Very intimate”. (Cite) have described the annotation process in detail.

The dataset is highly imbalanced, with majority of the labels in each language belonging to the lower spectrum of the scale as seen in Fig\ref{fig:overall}. Overall, 75\% of the samples in the dataset have a label less than or equal to 2.667.

The testing set additionally contains 4 unseen languages, Hindi, Korean, Arabic and Dutch.

\begin{figure}[h]
\includegraphics[width=\columnwidth]{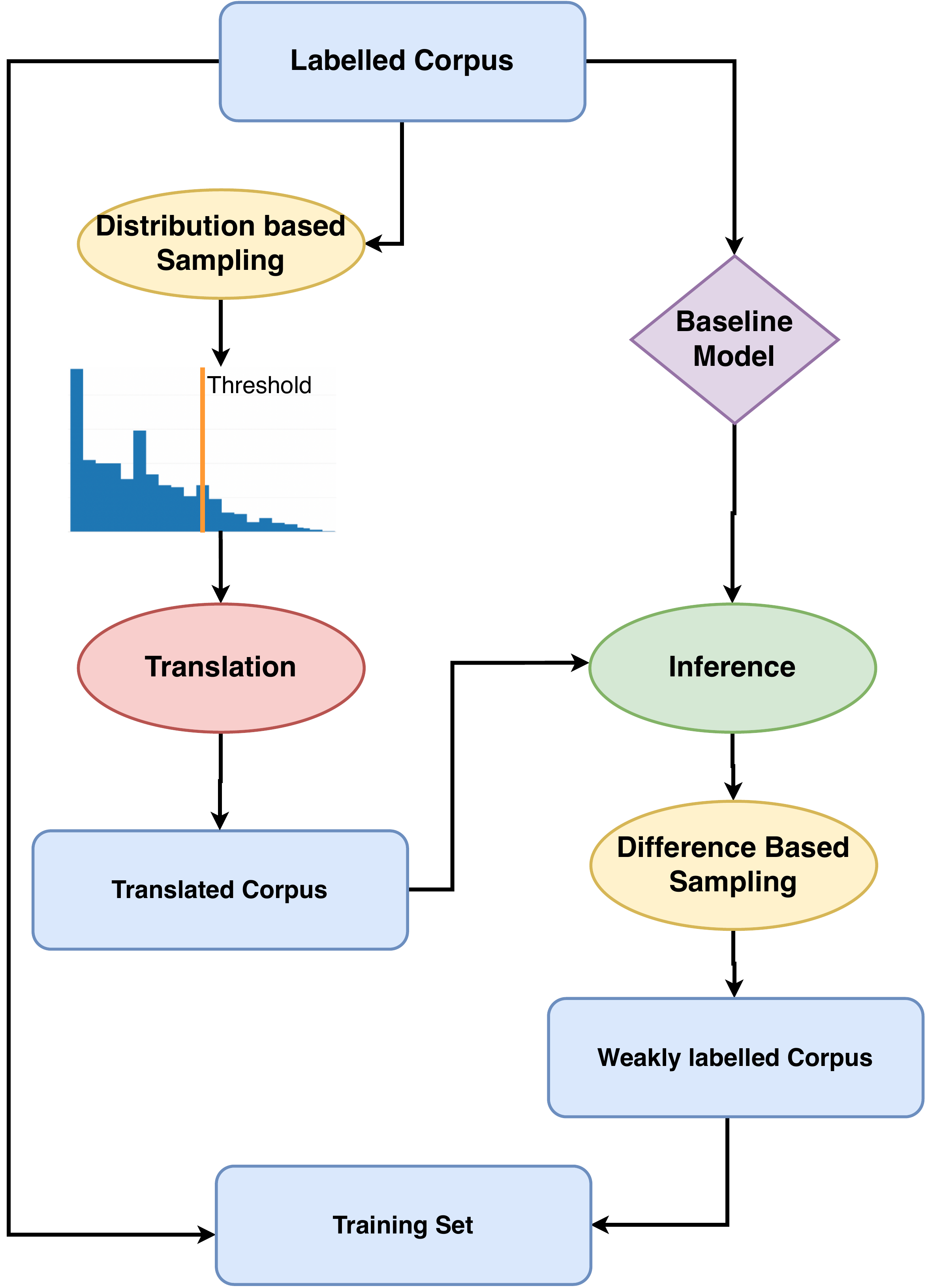}
\caption{Data Augmentation Flowchart}
\label{fig:dataaug1}
\end{figure}

\section{Methodology}
\subsection{Data Augmentation}
\begin{figure*}[h]
\includegraphics[width=2\columnwidth]{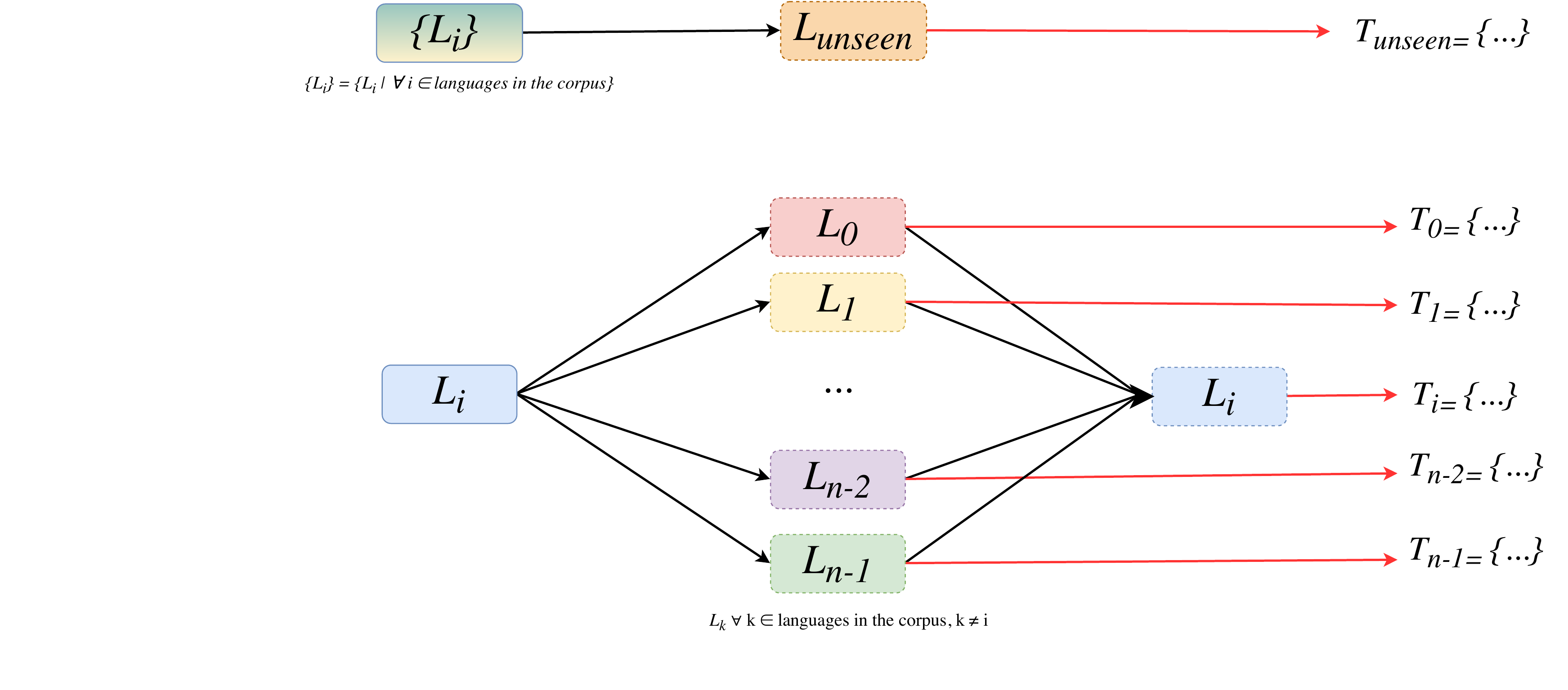}
\caption{Translation scheme}
\label{fig:dataaug2}
\end{figure*}

\begin{table*}[]
\centering
\begin{tabular}{lllllrrrl}
\hline
\textbf{Parameters} & \textbf{count}            & \textbf{mean}                & \textbf{std}                 & \textbf{min}          & \textbf{25\%} & \textbf{50\%} & \textbf{75\%} & \textbf{max}                 \\ \hline
\textbf{Value}      & \multicolumn{1}{r}{49774} & \multicolumn{1}{r}{0.62} & \multicolumn{1}{r}{0.51} & \multicolumn{1}{r}{0} & 0.23     & 0.47      & 0.86      & \multicolumn{1}{r}{3.550781}
\end{tabular}%
\caption{Analysis of the translated sentence set, specifically the difference during validation.}
\label{tab:diff2}
\end{table*}

As noted in section \ref{sec:data}, the data is highly imbalanced for the given labels. Moreover, since the task has 4 unseen languages, there is an additional need for data augmentation.
WADER performs data augmentation using the framework described in Fig \ref{fig:dataaug1}. The steps followed are described as follows:

\textbf{Distribution based Sampling:} Since the distribution of labels is skewed, and not all labels need augmentation, we perform a distribution based sampling to select candidate tweets for data augmentation. We fix a threshold $p$, and sample all tweets above the given threshold. We take the value of $p$ as 3.2, and less 

\textbf{Translation:} Data is augmented through translation and back translation. The translation scheme is described in Fig \ref{fig:dataaug2}.

For an unseen language, $L_{unseen}$, set of sampled sentences ${L_i}  \forall  i \in L$ are taken and translated to the target language, $L_{unseen}$. The translated sentences are appended to the set $T_{unseen}$.

For a seen language, say $L_i$, the language is translated to all other languages except the language itself, $L_k \forall k \in L, k \neq i$. The translated tweets are appended to their respective translated sets $T_k$, and they are translated back to the source language $L_i$, appended to the translated set $T_i$. 

Our final translated set has 49774 sentences.

\textbf{Label Validation:} We train a baseline model by finetuning a pretrained language model on the gold labelled data. This model is then used to infer on the concatenated translated corpus of seen and unseen languages. 
\begin{table}[h]
\centering
\begin{tabular}{lr}
\textbf{Difference} & \multicolumn{1}{l}{\textbf{Count}} \\ \hline
\textless{}= 0.1    & 5102                               \\
\textless{}= 0.2    & 10581                              \\
\textless{}= 0.3    & 16187                         
\end{tabular}
\caption{Count of sentences with chosen absolute difference threshold $\beta$ after label validation. }
\label{tab:diff}
\end{table}

\textbf{Difference Based Sampling:} We take the absolute difference between predicted and pre-assigned values (derived label from before a sentence was translated). We use this as a metric for quality of translations and pick appropriate thresholds to select sentences. Table \ref{tab:diff2} shows an analysis of the distribution of differences. The mean difference is 0.62, which is a below average translation quality since the resolution of labels is 0.1 in the dataset. However, 75\% of sentences have differences $\le$ 0.86 For our experiments, which means that coarse grain labels (differences of 1) are correctly assigned in most of the cases.

We define $\beta$ as the parameter which represents the difference threshold. We pick difference values of $\beta$ as 0.1, 0.2 and 0.3 in our experiments. Table \ref{tab:diff2} shows the count of sentences in each of these thresholds.

\subsection{Finetuning Pre-trained Language Models}
Finetuning pretrained language models has become a popular approach for natural language processing tasks in recent years. Transformer based \cite{transformers} Pretrained Language Models such as BERT \cite{bert}, GPT-2\cite{gpt2}, and RoBERTa\cite{roberta} are trained on massive amounts of text data, which allows them to capture complex linguistic patterns and structures. Finetuning involves taking a pretrained language model and further training it on a specific downstream task, such as sentiment analysis or question answering. This approach has been shown to achieve state-of-the-art performance on a wide range of natural language processing tasks, with significantly less data and computation needed compared to training a model from scratch. Finetuning also allows for the transfer of knowledge learned from a large, diverse set of data to a smaller, more specific task, making it a powerful technique for natural language processing research.

\begin{figure}[h]
\includegraphics[width=\columnwidth]{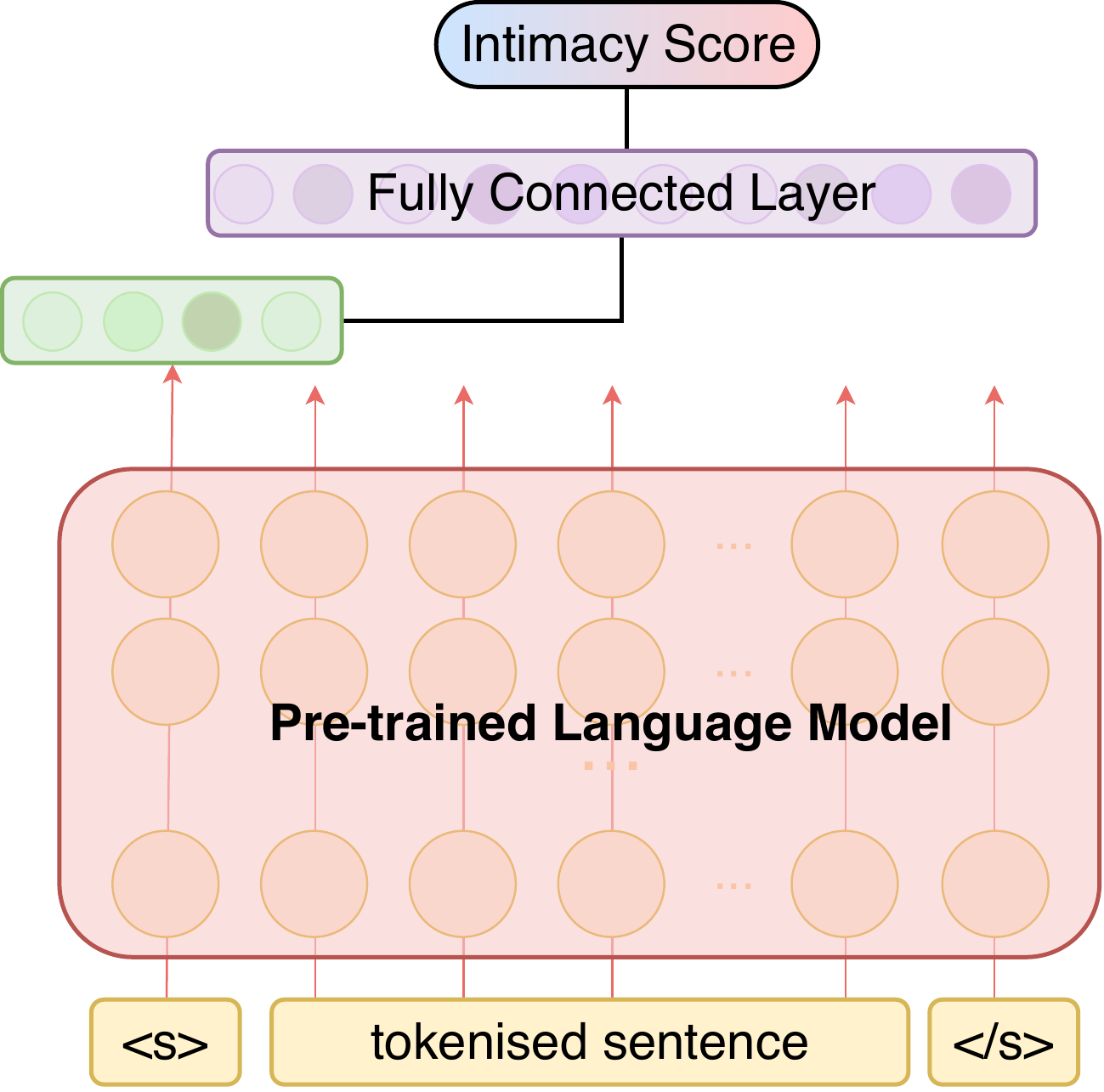}
\caption{Fine-tuning Architecture}
\label{fig:model}
\end{figure}

The pre-training models used in our system include:

\textbf{XLM RoBERTa:} XLM-RoBERTa \cite{xlmr} is a variation of the RoBERTa model that has been designed to handle multilingual natural language processing tasks. This model is pre-trained on a massive dataset of 2.5 terabytes of CommonCrawl data filtered for 100 different languages. By training on such a large and diverse dataset, XLM-RoBERTa is able to capture the linguistic nuances and patterns that are unique to different languages. The architecture of XLM-RoBERTa is based on the highly successful BERT model, but with key modifications to hyperparameters such as larger mini-batches and learning rates, allowing it to handle the additional complexity of multilingual data. XLM-RoBERTa has shown impressive results across a range of multilingual natural language processing tasks, demonstrating the power of pre-training on large, diverse datasets for building highly effective models. We use  implementation of the XLMR model from HuggingFace .

\textbf{XLNET:} XLNet \cite{xlnet} is a state-of-the-art natural language processing model that extends the Transformer-XL architecture and uses an innovative pre-training method. Unlike BERT, which corrupts input with masks and neglects dependencies between masked positions, XLNet is able to learn bidirectional contexts by maximizing the expected likelihood over all possible permutations of the factorization order. This allows XLNet to capture complex linguistic patterns and dependencies in the input sequence. XLNet also integrates ideas from Transformer-XL, which is currently the most advanced autoregressive model in use. With its autoregressive formulation, XLNet is able to overcome the limitations of BERT and achieve even better performance on a range of natural language processing tasks. 

For finetuning the pre-trained models, we add a single linear layer on top of the embeddings of the classification token \texttt{<s>} for XLM-RoBERTa and \texttt{[cls]}. Since this is a text regression task and the scores are in a limited, we apply a clamp function as the final activation function which clamps the scores in the range $[1,5]$. Fig \ref{fig:model} is a representation of our finetuning procedure.

\subsection{Ensembling}
\label{sec:ensemble}
We evaluate results on the test set using 6 models, XLM RoBERTa and XLNET trained on augmented sets with difference sampling parameter $\beta = 0.1, 0.2, 0.3$.

We choose 6 ensembles. The configurations of ensembles are defined in Table \ref{tab:ensemble}.

\begin{table}[h]
\centering
\resizebox{\columnwidth}{!}{%
\begin{tabular}{|c|ccc|ccc|}
\hline
 & \multicolumn{3}{c|}{\textbf{XLM RoBERTa}} & \multicolumn{3}{c|}{\textbf{XLNET}} \\ \cline{2-7} 
\multirow{-2}{*}{\textbf{Ensemble}} & \multicolumn{1}{c|}{\textbf{0.1}} & \multicolumn{1}{c|}{\textbf{0.2}} & \textbf{0.3} & \multicolumn{1}{c|}{\textbf{0.1}} & \multicolumn{1}{c|}{\textbf{0.2}} & \textbf{0.3} \\ \hline
\textbf{Ensemble 1} & \multicolumn{1}{c|}{\cellcolor{green}} & \multicolumn{1}{c|}{\cellcolor{green}} & \cellcolor{green} & \multicolumn{1}{c|}{\cellcolor{red}} & \multicolumn{1}{c|}{\cellcolor{red}} & \cellcolor{red} \\ \hline
\textbf{Ensemble 2} & \multicolumn{1}{c|}{\cellcolor{red}} & \multicolumn{1}{c|}{\cellcolor{red}} & \cellcolor{red} & \multicolumn{1}{c|}{\cellcolor{green}} & \multicolumn{1}{c|}{\cellcolor{green}} & \cellcolor{green} \\ \hline
\textbf{Ensemble 3} & \multicolumn{1}{c|}{\cellcolor{green}} & \multicolumn{1}{c|}{\cellcolor{red}} & \cellcolor{red} & \multicolumn{1}{c|}{\cellcolor{green}} & \multicolumn{1}{c|}{\cellcolor{red}} & \cellcolor{red} \\ \hline
\textbf{Ensemble 4} & \multicolumn{1}{c|}{\cellcolor{red}} & \multicolumn{1}{c|}{\cellcolor{green}} & \cellcolor{red} & \multicolumn{1}{c|}{\cellcolor{red}} & \multicolumn{1}{c|}{\cellcolor{green}} & \cellcolor{red} \\ \hline
\textbf{Ensemble 5} & \multicolumn{1}{c|}{\cellcolor{red}} & \multicolumn{1}{c|}{\cellcolor{red}} & \cellcolor{green} & \multicolumn{1}{c|}{\cellcolor{red}} & \multicolumn{1}{c|}{\cellcolor{red}} & \cellcolor{green} \\ \hline
\textbf{Ensemble 6} & \multicolumn{1}{c|}{\cellcolor{green}} & \multicolumn{1}{c|}{\cellcolor{green}} & \cellcolor{green} & \multicolumn{1}{c|}{\cellcolor{green}} & \multicolumn{1}{c|}{\cellcolor{green}} & \cellcolor{green} \\ \hline
\end{tabular}%
}
\caption{The configurations of the different chosen ensembles that we experimented with. The different choices are motivated by A) Model choice, B) Threshold of difference sampling $\beta$.}
\label{tab:ensemble}
\end{table}

Ensembling is done by taking the mean prediction of all the ensembled models.
\begin{table*}[t]
\centering
\resizebox{2\columnwidth}{!}{%
\begin{tabular}{lrrrrrrrrrrrrr}
\hline
\textbf{System} & \multicolumn{1}{l}{\textbf{Overall}} & \multicolumn{1}{l}{\textbf{Seen Langs.}} & \multicolumn{1}{l}{\textbf{Unseen Langs.}} & \multicolumn{1}{l}{\textbf{English}} & \multicolumn{1}{l}{\textbf{Spanish}} & \multicolumn{1}{l}{\textbf{Portuguese}} & \multicolumn{1}{l}{\textbf{Italian}} & \multicolumn{1}{l}{\textbf{French}} & \multicolumn{1}{l}{\textbf{Chinese}} & \multicolumn{1}{l}{\textbf{Hindi}} & \multicolumn{1}{l}{\textbf{Dutch}} & \multicolumn{1}{l}{\textbf{Korean}} & \multicolumn{1}{l}{\textbf{Arabic}} \\ \hline
\textbf{Baseline-XLM RoBERTa} & 0.52 & 0.65 & 0.35 & 0.60 & \textbf{0.69} & 0.60 & 0.64 & 0.60 & 0.70 & 0.19 & 0.59 & 0.37 & 0.42 \\
\textbf{0.1-XLM RoBERTa} & 0.52 & 0.66 & 0.34 & 0.61 & 0.66 & 0.60 & 0.67 & 0.63 & 0.72 & 0.19 & 0.59 & 0.35 & 0.48 \\
\textbf{0.2-XLM RoBERTa} & 0.52 & 0.67 & 0.33 & 0.63 & 0.66 & \textbf{0.61} & 0.67 & \textbf{0.64} & \textbf{0.72} & 0.19 & 0.60 & \textbf{0.38} & 0.49 \\
\textbf{0.3-XLM RoBERTa} & 0.53 & 0.66 & 0.35 & 0.63 & 0.67 & 0.60 & 0.67 & \textbf{0.64} & \textbf{0.72} & \textbf{0.20} & \textbf{0.61} & 0.43 & 0.50 \\
\textbf{Baseline-XLNET} & 0.38 & 0.51 & 0.22 & 0.62 & 0.61 & 0.42 & 0.47 & 0.47 & 0.24 & -0.08 & 0.37 & -0.03 & 0.05 \\
\textbf{0.1-XLNET} & 0.41 & 0.52 & 0.26 & \textbf{0.64} & 0.61 & 0.47 & 0.49 & 0.49 & 0.20 & -0.08 & 0.41 & -0.03 & 0.14 \\
\textbf{0.2-XLNET} & 0.41 & 0.51 & 0.29 & 0.61 & 0.58 & 0.43 & 0.50 & 0.51 & 0.24 & -0.05 & 0.45 & 0.08 & 0.22 \\
\textbf{0.3-XLNET} & 0.42 & 0.52 & 0.29 & 0.61 & 0.63 & 0.46 & 0.53 & 0.50 & 0.19 & -0.06 & 0.44 & 0.16 & 0.19 \\
\textbf{Ensemble-1} & 0.53 & 0.67 & 0.34 & 0.63 & 0.68 & \textbf{0.61} & \textbf{0.68} & \textbf{0.64} & 0.72 & 0.20 & 0.61 & 0.40 & \textbf{0.49} \\
\textbf{Ensemble-2} & 0.43 & 0.53 & 0.30 & 0.63 & 0.63 & 0.47 & 0.52 & 0.52 & 0.22 & -0.06 & 0.45 & 0.08 & 0.20 \\
\textbf{Ensemble-3} & 0.52 & 0.64 & 0.36 & \textbf{0.64} & 0.67 & 0.58 & 0.63 & 0.60 & 0.67 & 0.11 & 0.55 & 0.29 & 0.45 \\
\textbf{Ensemble-4} & 0.52 & 0.63 & 0.37 & 0.63 & 0.66 & 0.57 & 0.62 & 0.60 & 0.67 & 0.11 & 0.56 & 0.34 & 0.48 \\
\textbf{Ensemble-5} & 0.52 & 0.64 & \textbf{0.38} & \textbf{0.64} & \textbf{0.69} & 0.57 & 0.64 & 0.61 & 0.64 & 0.11 & 0.57 & 0.41 & 0.47 \\
\textbf{Ensemble-6} & \textbf{0.53} & \textbf{0.65} & 0.37 & \textbf{0.64} & 0.68 & 0.58 & 0.64 & 0.61 & 0.67 & 0.11 & 0.57 & 0.36 & 0.48 \\ \hline
\end{tabular}%
}
\caption{Pearson's R score of different system settings on the test set. $\beta-Model$ represents $Model$ finetuned on Gold labels + $\beta$ difference set.}
\label{tab:scores}
\end{table*}

\begin{table*}[h]
\centering
\resizebox{2\columnwidth}{!}{%
\begin{tabular}{llllllllllllll}
\hline
\textbf{Team} &
  \textbf{Overall} &
  \textbf{Seen Languages} &
  \textbf{Unseen Languages} &
  \textbf{English} &
  \textbf{Spanish} &
  \textbf{Portuguese} &
  \textbf{Italian} &
  \textbf{French} &
  \textbf{Chinese} &
  \textbf{Hindi} &
  \textbf{Dutch} &
  \textbf{Korean} &
  \textbf{Arabic} \\ \hline
\multicolumn{1}{r}{\textbf{WADER}} &
  \multicolumn{1}{r}{32} &
  \multicolumn{1}{r}{34} &
  \multicolumn{1}{r}{29} &
  \multicolumn{1}{r}{34} &
  \multicolumn{1}{r}{32} &
  \multicolumn{1}{r}{36} &
  \multicolumn{1}{r}{35} &
  \multicolumn{1}{r}{34} &
  \multicolumn{1}{r}{34} &
  \multicolumn{1}{r}{40} &
  \multicolumn{1}{r}{30} &
  \multicolumn{1}{r}{15} &
  \multicolumn{1}{r}{35} \\ \hline
\end{tabular}%
}
\caption{Rank achieved by our system in the shared task.}
\label{tab:rank}
\end{table*}

\section{Experimental Setup}
We use the original test and train set. Further, we take 15\% of the train set, sampled randomly from each language as our validation set.

We build our models using open source  available implementations of the XLM-RoBERTA and XLNET available on HuggingFace. We use \texttt{xlnet-base-cased} \texttt{xlm-roberta-base}\footnote{\url{https://huggingface.co/xlm-roberta-base}} and \footnote{\url{https://huggingface.co/xlnet-base-cased}}. We use Adam \cite{adam} as our optimiser. The size of the the embeddings are $D \in 768$ and the size of the linear layer is $D/2 \times 1$. The batch size is taken as 8 and the learning rate is 4e-5. We train the models for 2 epochs. Experiments are performed on Google Colab cloud GPU. Google Translate API has been used to perform translations. These hyperparameters are common for all system settings including our two baselines: 1) XLM RoBERTa finetuned on only Gold data, 2) XLNET finetuned on only Gold data.

Final submission is reported on Ensemble 6, configured as per the desciption in Section \ref{sec:ensemble}.
\section{Results and Discussion}
Table \ref{tab:scores} represents the scores achieved by our system in different experimental settings. The final submission for the competition is denoted by Ensemble 6. Table \ref{tab:rank} shows our rank under different categories of the shared task.

As we can observe from Table \ref{tab:scores}, WADER seems to improve on existing transformer baselines for all categories except one where it ties with an ensemble. 

\subsection{Comparison of Pre-trained Language Models:}
We observe a general trend that XLM RoBERTa performs better than XLNET on multilingual baselines in our experiments. This can be demonstrated by the fact that the XLNET Baseline outperforms XLM RoBERTa only on English. For all other languages, there is a significant margin in between performance of XLNET and XLM RoBERTa. For Hindi, and Korean which have non latin characters, performance of XLNET is even worse with a negative R coefficient. which has  This demonstrated the importance of multilingual pretraining.
\subsection{Comparison of Difference Sampling Threshold $\beta$:}
While lower values of $\beta (=0.1)$ give more accurate labelled sets, we observe that moderate values of $\beta (=0.1,0.2)$ outperform them. This is because, moderate values of $\beta$ allow for larger sized training corpuses, which would positively effect the performance of the models.  Moreover, moderate values $\beta$ include more number of low quality translations, due to a higher difference. We hypothesize that this would have a regularising effect by providing the model with diversity in the training set, and preventing it from overfitting on the training corpus.

\subsection{Discussion on Performance}
 We rank 32 overall, 34 on seen languages and 29 on unseen languages. The lower performance of our model can be understood by the following factors:
\begin{itemize}
    \item \textbf{Translation Quality:} The quality of translation is a key driver in WADER's performance. Lower quality translations would produce augmentations with noisy and unreliable labels. Translation quality is often dependant on the pair of languages in question. For languages such with a non latin script such as Hindi, translations are often of a lower quality which is also reflected in the results.
    
    \item \textbf{Overfitting:} By translating the data, while we increase linguistic diversity, most sentences would still be semantically similar, causing the model to overfit. This can further be seen by the fact that settings like 0.2, 0.3-XLM RoBERTA (where we can expect higher diversity from gold sentences due to higher differences) give the best performance for a lot of languages. Similarly, Ensemble 1 which preserves  data quality while also reaping the regularising benefit of ensembling performs quite well in the given setting. Another indication of overfitting is the the better rank of our model on unseen languages.

    \item \textbf{Word Sensitivity:} For a task like Intimacy Detection, specific vocabulary used is key to identify the intimacy level. Translations can lead to replacement of words which do not hold the same degree of influence in accounting for textual intimacy.
\end{itemize}

\section{Conclusion and Future Work}
This paper proposes a novel data augmentation framework, WADER, for text regression tasks that use weak-labeling strategies to solve the problems of data imbalance and data scarcity. We also provide a method for data augmentation in cross-lingual, zero-shot tasks. Our approach uses sampling techniques to mitigate bias in data and optimally select augmentation candidates. We benchmarked the performance of State-of-the-Art pre-trained multilingual language models XLM RoBERTa using WADER and achieved promising results. Our findings demonstrate the importance of data augmentation for mitigating data imbalance and scarcity in text regression tasks. This study's contributions provide a direction for future research in the field of computational linguistic and its applications to social information analysis. 

\bibliography{anthology,custom}
\bibliographystyle{acl_natbib}

\end{document}